%% file: clothformer.tex
\documentclass[10pt,twocolumn,letterpaper]{article}

\usepackage{cvpr}              

\usepackage{graphicx}
\usepackage{amsmath}
\usepackage{float}
\usepackage{amssymb}
\usepackage{booktabs}
\usepackage{makecell}
\usepackage{subfiles}
\usepackage[accsupp]{axessibility}  

\usepackage[pagebackref,breaklinks,colorlinks]{hyperref}

\usepackage[capitalize]{cleveref}
\crefname{section}{Sec.}{Secs.}
\Crefname{section}{Section}{Sections}
\Crefname{table}{Table}{Tables}
\crefname{table}{Tab.}{Tabs.}

\def\cvprPaperID{**} 
\def\confName{****}
\def\confYear{****}

\begin{document}
\renewcommand{\thefootnote}{\fnsymbol{footnote}}
\renewcommand{\paragraph}[1]{\vspace{1.25mm}\noindent\textbf{#1}}

\title{ClothFormer: Taming Video Virtual Try-on in All Module}

\author{Jianbin Jiang$^1$ \thanks{Work done in iQIYI Inc.} \hspace{20pt} Tan Wang$^2$  \hspace{20pt} He Yan$^2$\hspace{20pt} Junhui Liu$^2$ \\ 
{$^1 $} BIGO \hspace{20pt} {$^2 $} iQIYI Inc. \\
{\tt\small jiangjianbin@bigo.sg \hspace{20pt} \{wangtan, yanhe, liujunhui\}@qiyi.com} 
}

\maketitle

\begin{abstract}
The task of video virtual try-on aims to fit the target clothes to a person in the video with spatio-temporal consistency. Despite tremendous progress of image virtual try-on, they lead to inconsistency between frames when applied to videos. Limited work also explored the task of video-based virtual try-on but failed to produce visually pleasing and temporally coherent results. Moreover, there are two other key challenges: 1) how to generate accurate warping when occlusions appear in the clothing region; 2) how to generate clothes and non-target body parts (e.g. arms, neck) in harmony with the complicated background; To address them, we propose a novel video virtual try-on framework, ClothFormer, which successfully synthesizes realistic, harmonious, and spatio-temporal consistent results in complicated environment. In particular, ClothFormer involves three major modules. First, a two-stage anti-occlusion warping module that predicts an accurate dense flow mapping between the body regions and the clothing regions. Second, an appearance-flow tracking module utilizes ridge regression and optical flow correction to smooth the dense flow sequence and generate a temporally smooth warped clothing sequence. Third, a dual-stream transformer extracts and fuses clothing textures, person features, and environment information to generate realistic try-on videos. Through rigorous experiments, we demonstrate that our method highly surpasses the baselines in terms of synthesized video quality both qualitatively and quantitatively$\footnote[2]{Demos in video format are available at \url{https://github.com/luxiangju-PersonAI/ClothFormer}.}$.
\end{abstract}

\begin{figure}[t]
  \centering
   \includegraphics[width=1.0\linewidth]{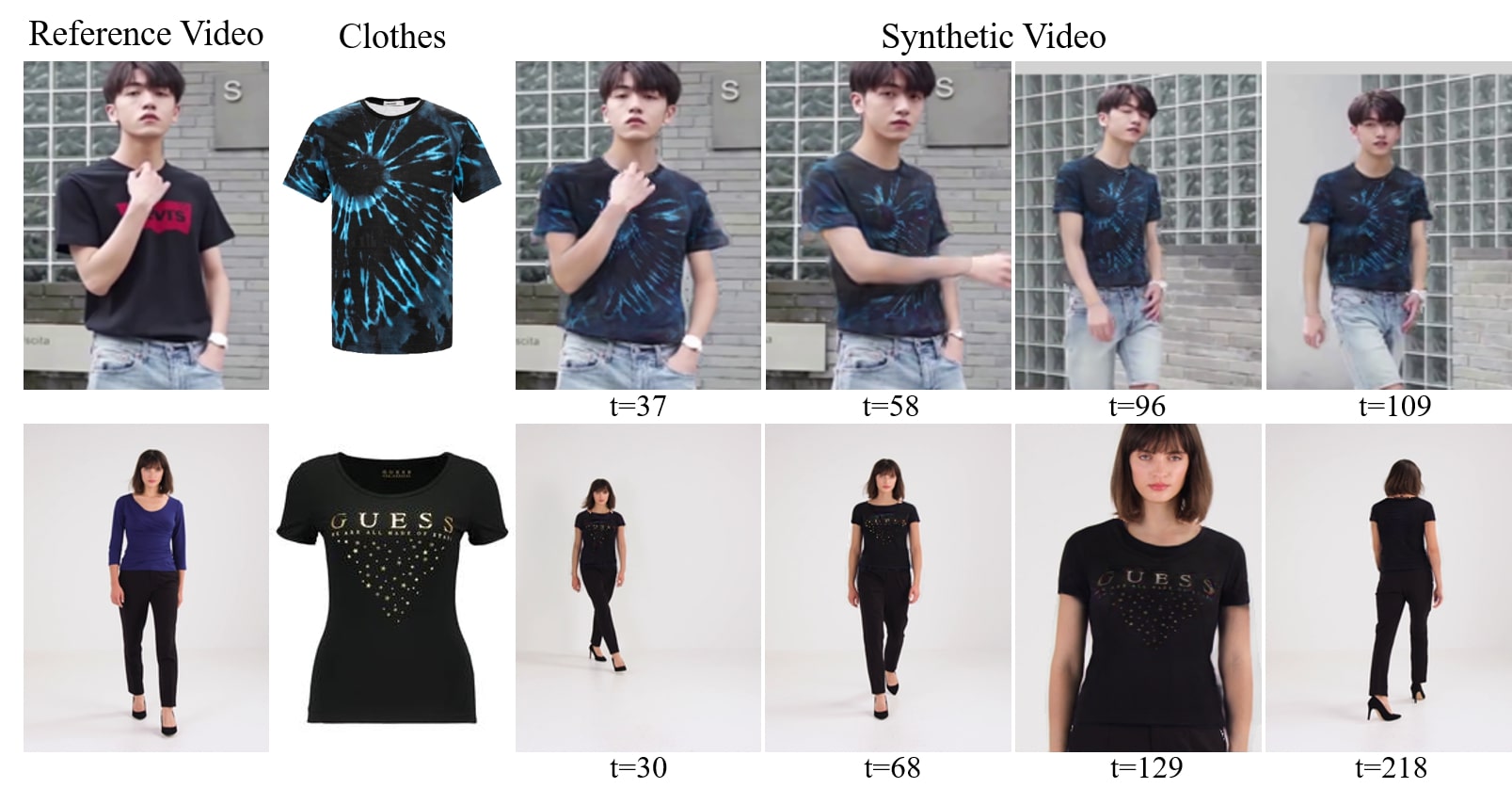}

   \caption{Examples of our video virtual try-on results on the VVT dataset and our dataset (first row). }
   \label{fig:cover}
\end{figure}

\section{Introduction}
\label{sec:intro}
The task of video virtual try-on aims to synthesize a coherent video that preserves the appearance of one target clothes and the original person's pose and body shape in the source video. This task has attracted much attention in recent years because of the prospects of its wide application in e-commerce and short video industry.

Previous virtual try-on methods usually focus on image-based operations~\cite{VITON,ACGPN,Cycle,clothflow,WUTON,CP-VTON,CP-Vton+,VITON-HD}. Among them, CP-VTON~\cite{CP-VTON} proposed a geometric matching module to learn the parameters of TPS transformation, which greatly improves the accuracy of deformation. WUTON~\cite{WUTON} and PFAFN~\cite{PFAFN} proposed parser-free methods to reduce the dependency of using accurate masks, and VITON-HD~\cite{VITON-HD} further increased the resolution of the generated image. These methods have achieved great success in different aspects. However, image virtual try-on is far from video virtual try-on in terms of immersion, and it often lead to inconsistent results between frames when image-based methods applied to videos.

There are a few attempts of designing video virtual try-on. FW-GAN~\cite{fwGan} is the first proposed method, which introduces the optical flow prediction module proposed in Video2Video~\cite{video2video} to warp the past frames to synthesize coherent future frames. Similarly, FashionMirror~\cite{fashionmirror} also predicts optical flow, however, it warps the past frames to future frames at the feature level instead of the pixel level, which enabled it can generates clothes in different views. MV-TON~\cite{mv-ton} further adds a memory refinement module to memorize the features of past frames. Although the above methods have made some progress in video spatio-temporal consistency, however, their generated frames are flickering and has a distance from achieving a spatio-temporally smooth video. We argue that there are two stems: on one hand, the above approaches only pay attention to try-on module while ignoring that the inputs deformed by the warping module are inconsistent. On the other hand, the images are synthesized in a frame by frame manner, which suffers from inconsistent attention results along spatio-temporal dimensions and often leads to blurriness and temporal artifacts in videos. Besides, the above approach is difficult to meet the demands in practical scenarios due to the presence of occlusions (\eg, hair, hands, bags) appearing in the clothing area. Lastly, the above methods are designed for simple datasets with pure background, like VVT~\cite{fwGan}. Thus they cannot deal with complex environment, nor to be in harmony with the natural background.

To address the challenges mentioned above, we propose a novel video virtual try-on framework, called ClothFormer. Firstly, inspired by VITON-HD~\cite{VITON-HD}, we introduce a clothing-agnostic person representation that eliminates clothing information thoroughly and preserves background and occlusion. Next, we novelly employ frame-level TPS-based warp method to predict and mask the occlusion region of the target clothes, then feed the processed target clothes to an appearance-flow-based methods to get a accurate and anti-occlusion dense flow mapping (appearance-flow) between the body regions and the clothing regions. Moreover, we enforce output to be spatio-temporally consistent in both warp module and try-on module. In warp module, we carry out two steps (ridge regression and optical flow correction) on appearance-flow sequence to produce the temporally smooth warped clothes sequence. In try-on module, we propose a \textbf{M}ulti-scale \textbf{P}atch-based \textbf{D}ual-stream \textbf{T}ransformer (\textbf{MPDT}) generator with multi-input and multi-output to synthesize the final realistic video based on the outputs from the previous stages, which simultaneously optimizes all the output frames in a single feed-forward process. Lastly, MPDT generator employs MPDT block to extract the clothes color and texture spatio-temporal features from warped clothing sequence and extract preserved person features and environmental information in agnostic-clothing sequence, which aims to generate clothes, non-target body(including arms, neck \etal) and fill the background in agnostic region. Then, the integration between the clothing features and background content features adopted in MPDT block to generate more harmonious results. To validate the performance of our framework, we collected a wild virtual try-on dataset with occlusion and complicated background for our research purpose.  Our experiments demonstrate that ClothFormer significantly outperforms the existing methods in generating videos, both quantitatively and qualitatively. 

Our contributions can be summarized as below:
\begin{itemize}
    \item  A novel warp module that combines the advantages of TPS-based methods and appearance-flow-based methods is designed to address the problem of inaccurate warp due to occlusions appear in clothing region.
    \item A tracking module based on ridge regression and optical flow correction are proposed to deforme a temporally smooth warped clothing sequence, which provides a prerequisite for the try-on module to generate coherent videos.
    \item The MPDT generator is designed carefully in the try-on module, which can extract and fuse clothing textures, person features and environment information to generate realistic try-on videos. To the best of our knowledge, this is the first time that transformer has been applied to the video virtual try-on.
\end{itemize}
\section{Related Work}
\label{sec:Related_Work}
\paragraph{Video processing and generation.} The video processing and generation techniques summarized here including video inpainting~\cite{deepflow, deep_video_inpainting}, video instance segmentation~\cite{Vistr}, video semantic segmentation~\cite{vseg1}, video super resolution~\cite{video_in_1}, video-to-video synthesis~\cite{video2video} and video virtual try-on~\cite{fwGan, mv-ton, fashionmirror}. 
These works share the common process of using a temporally consistent video sequence as input and harnessing temporal information to produce temporally smooth videos.
Inspired by previous works, FW-GAN~\cite{fwGan}, FashionMirror\cite{fashionmirror} and MV-TON~\cite{mv-ton} adopted various video processing methods to the task of virtual try-on and have proved their effectiveness. 
FW-GAN and FashionMirror predict optical ﬂow to warp the past synthesized frames at the pixel or feature level to generate subsequent frames, which was first proposed in vid2vid~\cite{video2video}. MV-TON proposes a memory refinement module to reconstruct spatio-temporal information, which was used in video semantic segmentation~\cite{vseg1, vseg2}.
However, these methods ignore that the input of try-on module, \ie, the warped clothing sequence is not smooth in the temporal dimension, which leads to blurriness and temporal inconsistency in videos. In contrast, we propose a tracking strategy based on optical flow and ridge regression to obtain a temporally consistent warp sequence as the input of the try-on module.

\paragraph{Vision Transformer.} The transformer~\cite{transformer} was first proposed for sequence-to-sequence machine translation task and have been recently adopted for computer vision tasks. The ViT~\cite{vit} was proposed by directly applying a pure transformer to sequences of 16x16 image patches for image classification tasks, which attains promising results compared to previous convolutional neural networks (CNNs).
DETR~\cite{DETR} built an end-to-end object detection method using bipartite-matching loss with transformer-based encoder and decoder, which largely simplifies the traditional detection pipeline~\cite{fast_rcnn} and achieves high performance and efficiency on par with CNN-based methods. 
Inspired by DETR, VisTR~\cite{Vistr} builts a simpler and faster video instance segmentation framework by using transformer. STTN~\cite{STTN} proposed a multi-scale patch-based joint Spatial-Temporal Transformer Network (STTN) for video inpainting and achieved state-of-the-art performance. These methods have proved the effectiveness of transformer in computer vision tasks. However, to the best of our knowledge, there is no previous study that successfully applies transformer in video virtual try-on. We find that the transformer can not only explore spatial correlation between patches by self-attention mechanisms but can also extract temporal correlation across multiple frames. Based on this idea, we propose a Dual-Stream Transformer for video virtual try-on.

\paragraph{Virtual Try-on.} Existing methods on virtual try-on can be classified as 3D-model-based approaches~\cite{3D1,3D2,3D3, 3D4, 3D5} and 2D-image-based ones~\cite{ACGPN, Cycle, VTNFP, clothflow, WUTON, CP-VTON, CP-Vton+, VITON-HD}. 3D-model-based approaches are not widely applicable due to the need of additional 3D measurements and high computational complexity, while 2D-image-based approaches are more broadly applicable.
VITON~\cite{VITON} designs a coarse-to-fine strategy which can seamlessly transfer a desired clothing item onto the corresponding region. The geometric matching module in CP-VTON~\cite{CP-VTON} well preserves the clothes identity in the generated image. WUTON~\cite{WUTON} and PFAFN~\cite{PFAFN} proposed parser-free methods to relieve the need of using accurate masks. VITON-HD~\cite{VITON-HD} synthesizes 1024×768 images by using ALIgnment Aware Segment (ALIAS) normalization and ALIAS generator. Compared with image virtual try-on, video virtual try-on is more practical and user-friendly. FW-GAN~\cite{fwGan} proposed a ﬂow-navigated warping GAN in video virtual try-on to generate coherent video stream. MV-TON~\cite{mv-ton} adopted memory refinement to improve the details by modeling the previously generated frames. Nevertheless, none of the above video virtual try-on methods can deal with the temporal consistency of warped input sequences. In this paper, we propose to use dual-stream transformer to process warped input sequences, by temporally aggregating and smoothing background and foreground information in spatio-temporal dimension to synthesize realistic video stream.


\begin{figure*}
  \centering
  \begin{subfigure}{01\linewidth}
    \includegraphics[width=1\linewidth]{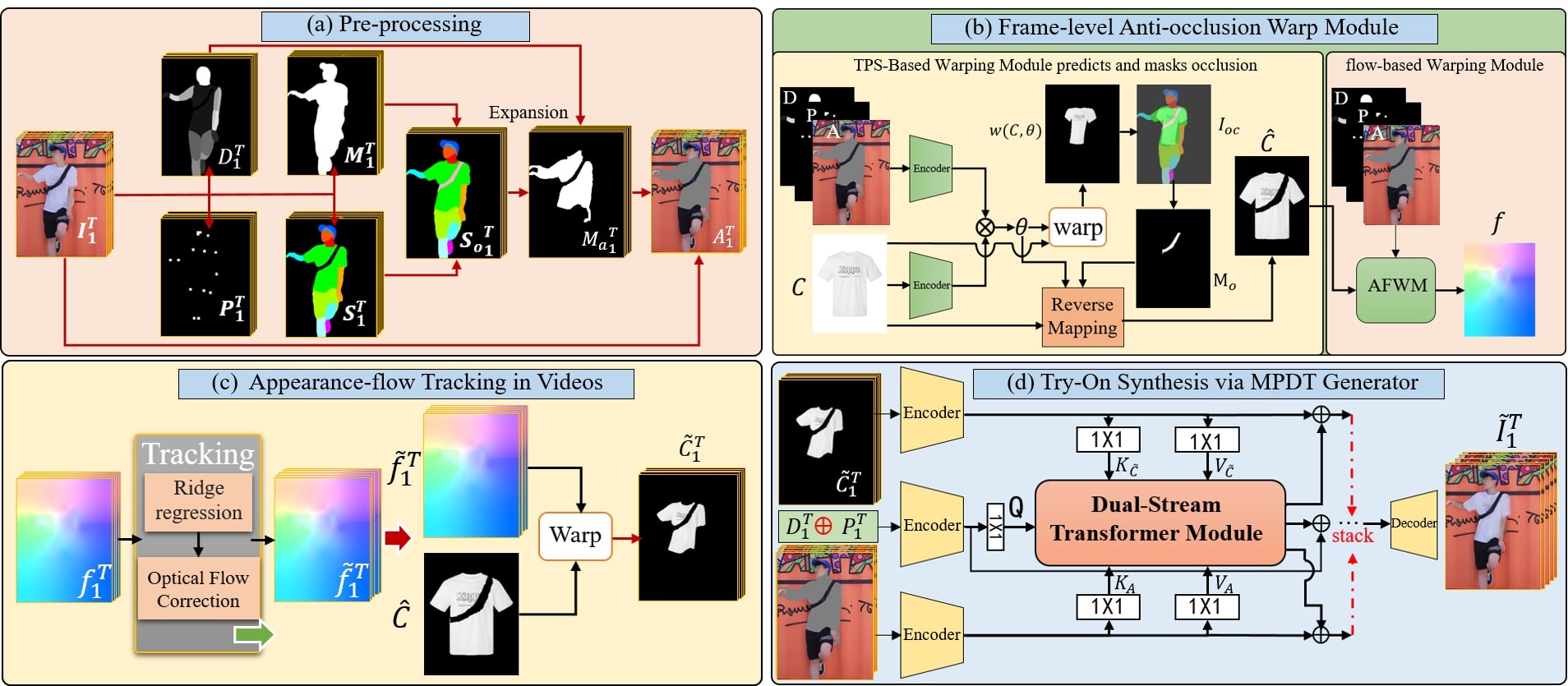}
    
    \label{fig:short-a}
  \end{subfigure}
  \caption{Framework of ClothFormer. (a) First, we obtain clothing-agnostic person image sequences $A_{1}^{T}$. (b) We predict t warped clothes $ W(C, \theta_{t}) $ by TPS-based warp method to infer an anti-occlusion target clothes $C$, then appearance-flow-based warp method is adopted to get an appearance flow $f$. (c) Appearance-flow tracking module based on ridge regression and optical flow correction is designed to get warped clothing sequence with spatio-temporally consistent. (d) Finally, MPDT generator synthesizes the final output video sequence $\Tilde{I}_{1}^{T} $ based on the outputs from the previous stages}
  \label{fig1}
  \vspace{-0.29cm}
\end{figure*}

\section{Proposed Method}

As described in~\cref{fig1}, given a target clothing image $C\in\mathbb{R}^{3\times H\times W}$ and a reference person video sequence  $I_{1}^{T}:=\{ I_{1},...,I_{T}\}\in\mathbb{R}^{3 \times H \times W}$, $H$ and $W$ denote height and width of the image, $T$ is the frame length of the sequence. 
ClothFormer aims to synthesis a realistic video sequence $ \Tilde{I}_{1}^{T}:=\{\Tilde I_{1},..., \Tilde I_{T} \}\in\mathbb{R}^{3 \times H \times W}$ that represents a person wearing the target clothes $C$ with spatio-temporal consistency, where the pose and body shape of $I_{1}^{T}$ as well as the color and texture of $C$ are preserved.
Training with sample triplets $( I_{1}^{T},C, \Tilde I_{1}^{T}) $ is straightforward but undesirable in practice~\cite{CP-VTON}. Instead, we use $( I_{1}^{T},C, I_{1}^{T})$ where the clothes $C$ already worn on the reference person video sequence $I_{1}^{T}$.

Since directly training on $ ( I_{1}^{T},C, I_{1}^{T}) $ harms model's generalization ability during inference, we construct an occlusion preserved clothing-agnostic person representation that eliminates the effects of source clothing in $ I_{1}^{T} $ clarified in \cref{31}. We combine the advantages of the TPS-based warping methods and appearance-flow-based methods to solve the problem of inaccurate warping due to occlusion, and add a tracking module to deform a temporally smooth warped clothing sequence, which are introduced in detail in \cref{32} and \cref{33}. Lastly, we propose a novel MPDT generator to synthesis realistic video in \cref{34}.

\subsection{Pre-processing}\label{31}
Inspired by~\cite{ACGPN, VITON-HD}, we propose constructing a clothing-agnostic video sequence $ A_{1}^{T} $ as inputs of warping and try-on modules, which preserves the person identity (\eg, face, hands and lower body) and eliminates clothing-gnostic regions with the occlusion preserved.
We obtain $ A_{1}^{T} $ by using the following four sequences, the segmentation map sequence $S_{1}^{T}:=\{S_{1},...,S_{T} \}$, the DensePose sequence $D_{1}^{T}:= \{D_{1},...,D_{T} \} $, the pose sequence $P_{1}^{T}:=\{P_{1},...,P_{T} \}$, and the matting sequence $M_{1}^{T}:= \{ M_{0},M_{1},...,M_{T}\}$. We use the pre-trained networks~\cite{parsing, densepose, openpose, matting} to generate these sequences. Specifically, as shown in \cref{fig1} (a), the $A_{1}^{T}$ is derived through masking out the clothing-gnostic regions of the $ I_{1}^{T} $ by utilizing clothing-gnostic mask $ {M_{a}}_{1}^{T} $, where $ {M_{a}}_{1}^{T} $ is expanded from arms, clothes and torso-skin region predicted in $S_{1}^{T}$ and then removing the hands and occlusion regions. Hands regions are predicted in $D_{1}^{T}$, the occlusion region is defined as the intersection of the foreground in $ M_{1}^{T} $ and the region predicted to be label zero in $ S_{1}^{T} $, where label zero in $ S_{1}^{T}$ denotes background or other items such as backpack strap on $ I_{1}^{T} $ in \cref{fig1} (a). Finally, the $ D_{1}^{T} $, $ P_{1}^{T} $ and $ A_{1}^{T} $ are the inputs of warping module and try-on module to generate temporally consistent video $ \Tilde{I}_{1}^{T} $.

\subsection{Frame-level Anti-occlusion Warp Module}\label{32} 
The existing warping methods cannot handle occlusions that appear in clothing regions. Most of them utilize appearance-flow-based methods or TPS-based methods to deform clothing images. The appearance-flow-based methods are very sensitive to occlusions, when there are occlusions (\eg, hair, arms, bags), the pixel-squeeze phenomenon is likely to occur as shown in the third column of \cref{anti1}. TPS-based methods can handle partial occlusions by estimating grid mapping to deform clothing images. However, it often leads to the misalignment between the warped clothes and the body~\cite{VITON-HD}. To address these issues, we propose a two-stage anti-occlusion strategy.

In the first stage, we adopt a TPS-based warping module. As shown in \cref{fig1}(b), we use ($A_{t}, D_{t}, P_{t}$) and the reference clothes $C$ as inputs at frame $t$. The TPS-based warping loss is formulated as:
\begin{equation}
  L^{TPS\textrm{-warp}}_{t}=\left\|I^{C}_{t}-W(C, \theta_{t})\right\|_1+\lambda^{s d c}_{t} L^{s d c}_{t}
  \label{eq:important}
\end{equation}
where $ I^{C}_{t} $ is the target clothes extracted from $I_{t}$ at frame $t$, $ W(C, \theta_{t}) $ is the warped clothes that deforms $C$ using $ \theta_{t} $, and $ \lambda^{s d c}_{t} $ is the hyper-parameter for second-order difference constraint $ L^{s d c}_{t} $ ~\cite{ACGPN}.

Afterward, we define the region where  $ W(C, \theta_{t}) $ overlaps with the location of occlusion in ${ S_{o}}_{t} $ as occluded region of $ W(C, \theta_{t}) $, as $ I_{oc} $ shown in \cref{fig1}(b), and the occluded area shown as $M_{o}$. Then we get the target clothes $ \hat{C}_{t} $ that have masked the occluded area by using $ \theta_{t} $ to reverse mapping. Finally, AFWM ~\cite{PFAFN} network adopted to learn dense flow mapping between the body regions and the clothing regions (appearance-flow $f_{t}$) with ($A_{t}, D_{t}, P_{t}$) and the $ \hat{C}_{t} $ as inputs. The appearance-flow $f_{t}$ is optimized as follows:
    \begin{equation}
  L^{flow-\text { warp }}_{t}=\left\|I^{C}_{t}-W( \hat{C}_{t}, f_{t})\right\|+\lambda^{s e c}_{t} L^{s e c}_{t}
  \label{flow loss}
\end{equation}
where $ W( \hat{C}_{t}, f_{t}) $ is the warped clothes that deforms $ \hat{C}_{t} $ using learned appearance-flow $ f_{t} $ at frame $t$, and $ \lambda^{s e c}_{t} $ is the hyper-parameter for second-order smooth constraint $ L^{s e c}_{t} $ ~\cite{PFAFN}.

\subsection{Appearance-flow Tracking in Videos}\label{33}
For video virtual try-on, the texture and the color of the synthesized clothes are mainly related to the input warped clothing sequence. Previous works~\cite{mv-ton, fwGan, fashionmirror} only focused on the temporal consistency of the try-on module but ignored the temporal consistency of the warped clothing sequence. Instead, we propose a appearance-flow tracking module to produce a temporally smooth warped clothing sequence by tracking the appearance-flow learned in \cref{32} as shown in \cref{fig1}(c).
The appearance-flow learned in warping module represents the coordinates of the pixels in the input clothes $ \hat{C}_{t} $ mapped to which position in the warped clothes $ W( \hat{C}_{t}, f_{t}) $, from this point of view the appearance-flow-based warp module is similar to facial landmark detection task~\cite{lmk}, the $ \hat{C}_{t} $ and $ W( \hat{C}_{t}, f_{t}) $ are analogous to aligned face image and the face image with different poses in facial landmark detection task. Inspired by ~\cite{tracking1}~\cite{tracking2}, we first reshape the $ f_{t} $ with the height $ H $ and width $ W $ into 1-dimension vectors $ f_{t}^{1D} $ of length $ W \times H $, and track it by developing a post-processing algorithm based on ridge regression which exploits correlation among adjacent flow to achieve a temporally smooth results. The appearance-flow optimized as:
  \begin{equation}
  \hat{f}_{t}^{1D}=X\left(X^{T} X-\mu I\right) X^{T} f_{t}^{1D}
  \label{ridge}
\end{equation}
where $ f_{t} $ is the appearance-flow of frame $ t $ and  $ 1\leq t \leq T$, $ I $ is the unit matrix, $ \mu $ is the hyper parameter and $ X $ is the feature matrix detailed in the supplementary.

In addition, the motion information in the clothing area of input person sequence $ I_{1}^{T} $ is crucial because it is not only related to the human pose but also related to the environmental factors such as wind. Therefore, we used the optical flow~\cite{flownet} in the clothing area to correct the $ \hat{f}_{t} $, denoted as:

\begin{equation}
 \Tilde{f}_{t}=\left\{
\begin{array}{rcl}
\frac{\hat{f}_{t}+\widetilde{w}_{t-1}\left(\hat{f}_{t-1}\right)}{2},       &      &  	\delta_{t} \leq \varepsilon \quad and \quad \hat{f}_{t} \in \Omega \\
\tilde{f}_{t}=\hat{f}_{t} \quad,     &      & 	\delta_{t}>\varepsilon \quad or \quad \ \ \  \hat{f}_{t} \notin \Omega
\end{array} \right. 
 \label{flow_res}
\end{equation}
where $ \hat{f}_{t}  $ is the tracking result of \cref{ridge}, $\widetilde{w}_{t-1}$ is the estimated optical flow from $ I_{t-1} $ to $ I_{t} $ of input person sequence. $ \Omega $ denotes 
the intersection of the clothing region on $ I_{t} $ and the warped clothing region of $ W( \hat{C}_{t}, \hat f_{t}) $. By $ \widetilde{w}_{t-1}(\hat{f}_{t-1}) $, we warp $ \hat{f}_{t-1} $ based on $\widetilde{w}_{t-1}$. $\delta_{t} $ defined as $ \left\|\hat{f}_{t}-\widetilde{w}_{t-1}\left(\hat{f}_{t-1}\right)\right\| $ and we set threshold value $ \varepsilon $ to 0.05. 

Finally, the warped clothing sequence with temporal consistency and anti-occlusion obtained as :
  \begin{equation}
  \Tilde{C}_{1}^{T} = W(\hat C_{1}^{T}, \Tilde{f}_{1}^{T} )
  \label{final_warp}
\end{equation}
\subsection{Try-On Synthesis via MPDT Generator}\label{34}

\begin{figure*}[t]
  \centering
  \begin{subfigure}{01\linewidth}
    \includegraphics[width=1\linewidth]{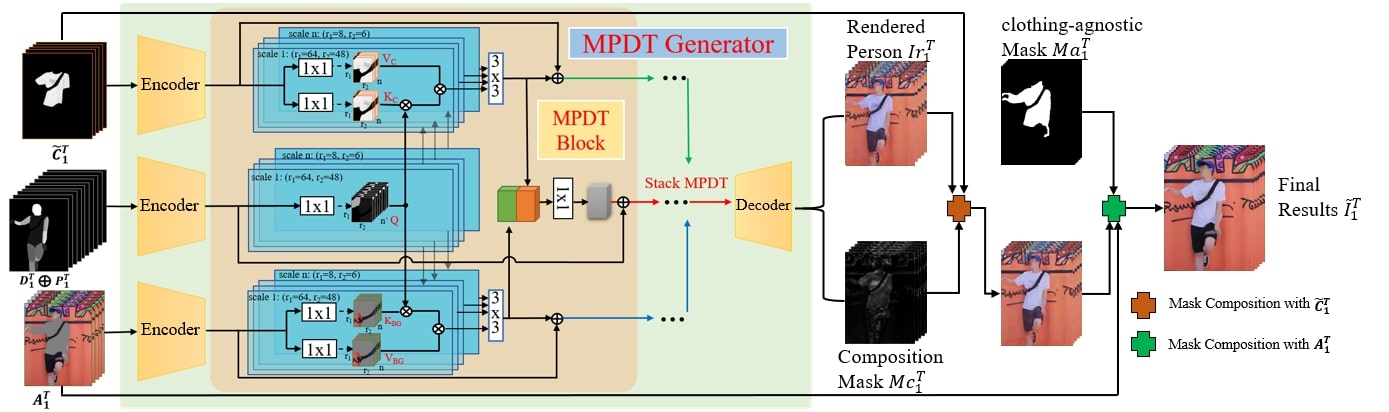}
    
    \label{figtyon}
  \end{subfigure}
  \caption{Illustration of the MPDT generator, including three frame-level encoders, stacked \textbf{MPDT} block, and a frame-level decoder. MPDT block is the core module, on one hand, MPDT block search and extract the texture and the color content from warped clothing sequence $\Tilde{C}_{1}^{T} $, on the other hand MPDT block borrow the environment and person identity information from clothing-agnostic person image sequence $A_{1}^{T}$ to synthesis body, fill masked background and make the generated clothes more harmonious with the environment. }
  \label{tyon1}
  \vspace{-0.4cm}
\end{figure*}

\def\model{MPDT\xspace}

In the try-on module, we propose the \model generator to synthesize realistic video sequences based on the outputs of the previous stages. \model generator can deal with the blurriness and temporal artifacts from which previous methods~\cite{mv-ton, fwGan} suffer a lot.

As shown in \cref{tyon1}, there are three inputs for \model generator: (1) warped clothing sequence $ \Tilde{C}_{1}^{T} $ generated in~\cref{33}; (2) person-shape sequence $ D_{1}^{T} \oplus P_{1}^{T} $ concatenated by DensePose sequence $ D_{1}^{T} $ and the pose sequence $ P_{1}^{T} $ generated in ~\cref{31}; (3) the clothing-agnostic sequence $ A_{1}^{T} $. For model architecture, \model generator consists of three components: three frame-level encoders, \model block, and a frame-level decoder. The \model block based on Transformer is the core component which aims to aggregate spatio-temporal features from the warped clothing stream and the clothing-agnostic stream. 


\paragraph{Embedding.} The embedding of query and key-value pairs play a crucial role in Transformer. For \model block, there are two sets of key-value pairs embedding are designed and a set of queries:
\begin{gather}
\begin{aligned}
q_{t}&=\textrm{conv}_{q}\left(p_{t}\right) \\
\left(k_{t}^{C}, v_{t}^{C}\right)&=\left(\textrm{conv}_{k^C}\left(C_{t}\right), \textrm{conv}_{v^C}\left(C_{t}\right)\right) \\
\left(k_{t}^{A}, v_{t}^{A}\right)&=\left(\textrm{conv}_{k^A}\left(A_{t}\right), \textrm{conv}_{v^A}\left(A_{t}\right)\right)
\end{aligned}
\label{embed}
\end{gather}
where $ 1\leq t\leq T $, $\textrm{conv}$ denotes the 1x1 2D convolutions. 

\paragraph{Dual-stream Spatio-temporal Attention.} Inspired by STTN~\cite{STTN}, we conduct attention among all spatio-temporal patches in the input video sequence. Each frame is segmented into patches of size $r_{1}\times r_{2}\times cn$ with the channel size $cn$, thus there are $N=T\times h/r_{1}\times w/r_{2}$ patches for the entire sequence totally. Before calculating patch-wise similarities, the query and key patches are reshaped into 1D vectors. The attention weights between query patches and clothing key-value patches denote as:

\begin{equation}
\begin{aligned}
\alpha_{i,j}^{C}= \begin{cases}\operatorname{softmax}_{j}\left(\frac{p_{i}^{q} \cdot\left(p_{j}^{k^{C}}\right)^{T}}{\sqrt{r_{1} \times r_{2} \times cn}}\right), & p_{j}^{k^{C}} \in \Omega^{C} \\ 0, & p_{j}^{k^{C}} \notin \Omega^{C}\end{cases}
  \label{tryon_2}
 \end{aligned}
\end{equation}

\begin{equation}
\begin{aligned}
Att_{i}^{C}= \sum_{j=1}^{N}\alpha_{i,j}^{C}.p_{j}^{v^{C}}
\label{tryon_11}
\end{aligned}
\end{equation}
where $ p_{i}^{q} $ denotes $i^{\textrm{th}}$ query patch. $ p_{j}^{k^{C}} $ and $ p_{j}^{v^{C}} $ denote the $j^{\textrm{th}}$ key-value patches. $ \Omega^{C} $ denotes the visible clothing regions. Similarly, the attention value $ Att_{i}^{A} $ between query patches and clothing-agnostic patches is calculated in the same way as \cref{tryon_11}. 
Then attention is applied in a multi-head manner and the attention values from different heads are concatenated as $A t t^{C}$ and $A t t^{A}$. We fuse these two streams through concatenation followed by a 1x1 convolution:

\begin{equation}
o=\left(A t t^{C} \oplus A t t^{A}\right) W_{1}+b_{1}
  \label{tryon_3}
\end{equation}
where $ \oplus $ denotes the concatenation, $ W_{1} $ and $ b_{1} $ are learnable parameters of 1x1 convolutions. 

The result $ o $ and query $q$ (added by residual connection) in current \model block will serve as the query of the next block. Then, a frame-level decoder simultaneously renders person image sequence ${I_R}_{1}^{T}$ and predicts composition mask sequence $ {M_{C}}_{1}^{T} $. We fuse ${I_R}_{1}^{T}$ and the warped clothing sequence $ \Tilde{C}_{1}^{T} $ by using $ {M_{C}}_{1}^{T} $ to enhance the texture details of the generated clothes, it is essential for $ \Tilde{C}_{1}^{T} $ to be temporally smooth.
\begin{equation}
{I_\textrm{masked}}_{1}^{T}= {M_{C}}_{1}^{T} \odot \Tilde{C}_{1}^{T}+(1-{M_{C}}_{1}^{T}) \odot {I_R}_{1}^{T}
  \label{fusion2}
\end{equation}

To reconstruct complex background and focus on the task of virtual try-on, the clothing-agnostic image and ${I_\textrm{masked}}_{1}^{T} $ are then fused together using the clothing-gnostic mask sequence $ {M_{a}}_{1}^{T} $ defined in \cref{31} to synthesize the final output $\Tilde{I}_{1}^{T} $:

\begin{equation}
\Tilde{I}_{1}^{T}= (1-{M_{a}}_{1}^{T})  \odot {I_\textrm{masked}}_{1}^{T} + \\{M_{a}}_{1}^{T} \odot A_{1}^{T}
  \label{fusion}
\end{equation}

Finally, we use spatio-temporal losses to train \model. In spatial dimension, we include $l_1$ loss and perceptual loss~\cite{perceptual} to ensure per-pixel reconstruction accuracy. In temporal dimension, we use a Temporal PatchGAN (TPGAN)~\cite{Tpatch, TP2} as the discriminator to improve the temporal consistency in generated video. The overall objective function is as:
\begin{equation}
L_{\textrm {try-on}}=\lambda_{1} L_{l 1}^{\text {whole}}+\lambda_{2} L_{l 1}^{clothes}+\lambda_{3} L_{\textrm {perc}}+\lambda_{4} L_{\textrm {TPGAN}}
  \label{tryon_loss}
\end{equation}
where $ L_{l1}^{whole} $ denotes the L1 loss of the whole image,  $ L_{l1}^{clothes} $ denotes the L1 loss in clothing regions. $ L_{TPGAN} $ is the adversarial loss. $ \lambda_{i}, i\in\{1,2,3,4\} $ are hyper-parameters.

\begin{table}
\setlength\tabcolsep{2pt}%
  \centering
  \begin{tabular}{l|ccccc}
    \toprule
    Method & Dataset & SSIM & LPIPS & \makecell[c]{VFID \\ 
    \footnotesize I3D } & \makecell[c]{VFID \\ 
    \footnotesize ResNeXt101 } \\
    \midrule
    CP-VTON\cite{CP-VTON} & VVT & 0.459 & 0.535 & 6.361 & 12.10\\
    ACGPN\cite{ACGPN} & VVT & 0.853 & 0.178 & 9.777 & 11.98 \\
    PBAFN\cite{PFAFN} & VVT & 0.870 & 0.157 & 4.516 & 8.690 \\
    FW-GAN\cite{fwGan} & VVT & 0.675 & 0.283 & 8.019 & 12.15 \\
    MVTON\cite{mv-ton} & VVT & 0.853 & 0.233 & 8.367 & 9.702 \\
    \textbf{ClothFormer} & VVT & \textbf{0.921} & \textbf{0.081} & \textbf{3.967} & \textbf{5.048} \\
    \midrule
    CP-VTON\cite{CP-VTON} & ours & 0.682 & 0.299 & 13.11 & 31.19 \\
    ACGPN\cite{ACGPN} & ours & 0.786 & 0.243 & 16.21 & 32.54 \\
    PBAFN\cite{PFAFN} & ours & 0.841 & 0.188 & 11.15 & 28.62 \\
    FW-GAN\cite{PFAFN} & ours & 0.705 & 0.344 & 13.71 & 28.31 \\
    CP-VTON*\cite{CP-VTON} & ours & 0.929 & 0.068 & 7.463 & 11.30 \\
    ACGPN*\cite{ACGPN} & ours & 0.936 & 0.066 & 10.89 & 13.91 \\
    PBAFN*\cite{PFAFN} & ours & 0.932 & 0.066 & 6.132 & 10.88 \\
    \textbf{ClothFormer$\dagger$} & ours & 0.953 & 0.047 & 5.071 & 9.018 \\
    \textbf{ClothFormer$\divideontimes$} & ours & \textbf{0.959} & 0.042 & 5.140 & 9.394 \\
    \textbf{ClothFormer$\diamond$} & ours & 0.949 & 0.050 & 5.653 & 9.721 \\
    \textbf{ClothFormer-tiny} & ours & 0.955 & 0.045 & 5.208 & 9.153 \\
    \textbf{ClothFormer} & ours & 0.958 & \textbf{0.042} & \textbf{5.024} & \textbf{8.971} \\
    \bottomrule
  \end{tabular}
  \caption{Comparison with previous methods on the VVT dataset and our new collected dataset. For SSIM, the higher is the better. For LPIPS and VFID, the lower is the better. ClothFormer$\dagger$, ClothFormer$\divideontimes$, ClothFormer$\diamond$ and ClothFormer-tiny are ClothFormer variants for ablation study.}
  \label{tab:example}
\end{table}

\section{Experiments}
\begin{figure*}
  \centering
  \setlength{\abovecaptionskip}{0.cm}
    \includegraphics[width=1\linewidth]{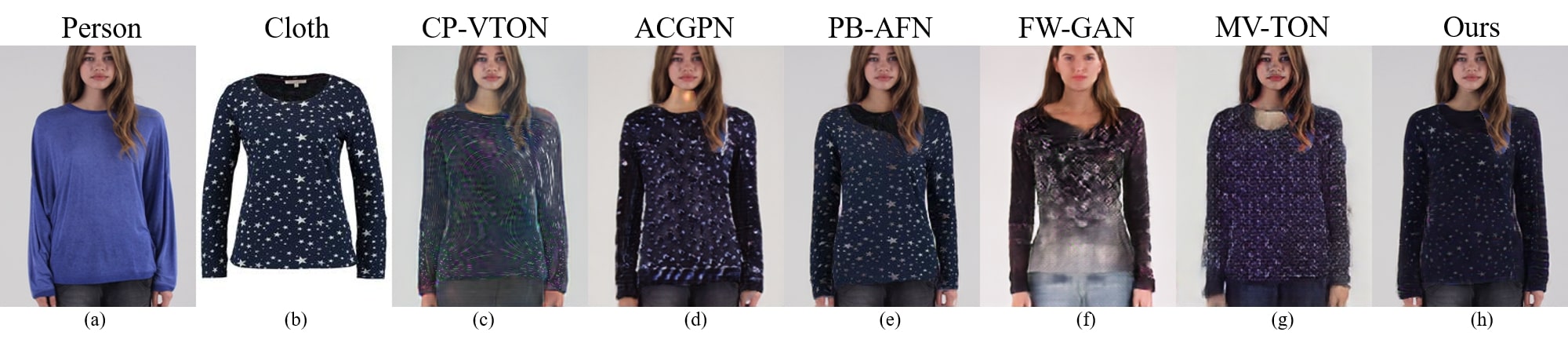}
    
    \label{fig:short-vvt}
  \caption{Visual comparison with the baseline methods on the VVT dataset. (a) is the reference person, (b) is the target clothes. ClothFormer produces a more temporally consistent video output and clearly preserve the details of the target clothes. } 
  \label{figvvt}

\end{figure*}
\begin{figure*}
  \centering
  \setlength{\abovecaptionskip}{0.cm}
  \begin{subfigure}{01\linewidth}
    \includegraphics[width=1\linewidth]{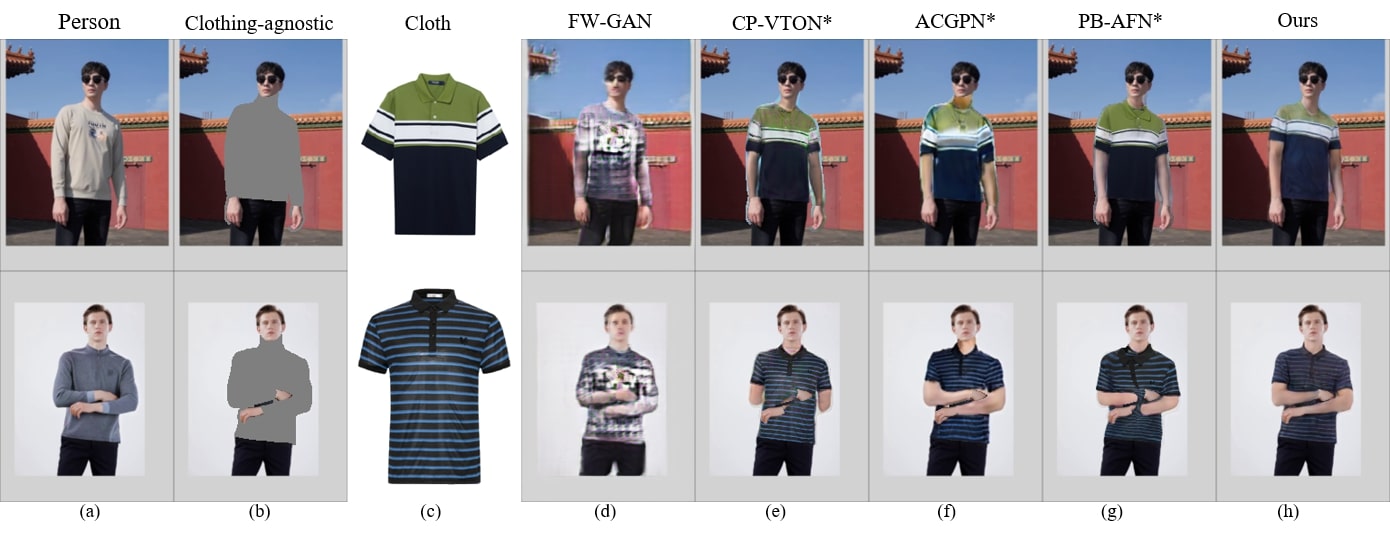}
    
    \label{fig:short-our}
  \end{subfigure}
  \caption{Visual compare with the baseline methods on our dataset, (b) is the clothing-agnostic images for composition as \cref{fusion} when training CP-VTON*, ACGPN*, PB-AFN* and ClothFormer. The first row shows ClothFormer generates a more harmonious results, the second row shows ClothFormer generates more satisfactory results when cross-arms appear in clothing region. } 
  \label{figour}
  \vspace{-0.4cm}
\end{figure*}
\subsection{Experiment Setup}\label{datasets} 
\paragraph{Datasets.} Experiments are conducted on the VVT~\cite{fwGan} dataset and our collected dataset. The VVT dataset contains 791 videos with the resolution of 192×256. The train and test set contain 159,170 and 30,931 frames respectively. However, in the VVT dataset, the backgrounds are primarily white, and the human poses are monotonous and simple. 
In contrast, we collected a wild virtual try-on dataset with complex environment, complicated poses and occlusions, which contains 3995 videos. The train and test set contain 179,965 and 25,710 frames respectively. In addition, every video in our dataset is divided into several coherent sub videos by adopting shot transition detection~\cite{shot_det}. 


\paragraph{Training and Testing.} We train the warp module and the try-on module separately and combine them to generate the try-on image eventually. The paired setting used to train the modules in the training process. 
In the testing process, we use the pairs of a person and a clothing image to evaluate a paired setting and shuffle the clothing images for an unpaired setting as in previous methods~\cite{VITON-HD}~\cite{ACGPN}. Moreover, the videos in VVT dataset are divided into coherent sub videos to train and test methods.

\subsection{Qualitative Analysis}\label{QA} 
We first compare proposed methods with video-based method FW-GAN~\cite{fwGan}, MV-TON\cite{mv-ton} and image-based methods CP-VTON~\cite{CP-VTON}, ACGPN~\cite{ACGPN} and PB-AFN~\cite{PFAFN} in VVT dataset for a more comprehensive experimental comparison. Moreover, to verify the superior performance of our model in complex environments and with occlusions appear in the person images, ClothFormer compared with all the above methods except MVTON~\cite{mv-ton} in our dataset for whose testing code and pretrained model on VVT dataset are available while the training code is not. (\textbf{Notes:} It is necessary to watch the videos to compare the qualitative results at urls or supplementary materials). 

\cref{figvvt} shows some qualitative results on the VVT dataset. Clothes generated by CP-VTON, ACGPN, FW-GAN, MVTON show many visual artifacts, including blurring and cluttered texture. Although each frame synthesized by PB-AFN is photorealistic, the resulting video lacks temporal coherence. Specifically, the texture of the clothes generated by PB-AFN is flickering irregularly even when the person keeps still. Compared with the baseline methods, ClothFomer produces a temporally consistent video output and preserves the details of the target clothes. 

To compare the performance of ClothFormer and the baseline methods in complex environments and the person with occlusions appearing in the clothing region, we conduct experiments in our new collected dataset. We first trained baseline methods CP-VTON, ACGPN and PBAFN according to their original setting. However, unlike the VVT dataset, the baseline methods could reconstruct the white background; the background of the reference person in our dataset is too complicated to reconstruct for these methods. Hence, for a fair comparison, like ClothFormer, we fuse the clothing-agnostic region and the output of these method during training process, which  are denoted as CP-VTON*, ACGPN* and PBAFN* respectively. 
From the sample results in the first row of \cref{figour}, it can be easily observed that our method achieves much better visual consistency compared to the baseline methods. For one thing, the clothes and body generated by baseline methods look unnatural in complex environments while the clothes and body generated by ClothFomer are in perfect harmony with the complex background. For another thing, ClothFormer can fill the region around generated body and clothes with plausible content in a video while the baseline methods failed. From the sample results in the second row of \cref{figour}, when occlusions like cross-arms appear in the person images, PB-AFN generates unnatural results with pixel-squeeze phenomenon, CP-VTON fails to generate arms and ACGPN generates fake and flickering arms. In comparison, our ClothFormer can warp the clothes to the target person accurately even when occlusion appears. The two rows of FW-GAN results do not even generate a humanoid , since FW-GAN only uses the RGB information of the first frame and the pose information of the subsequent frames to generate video, it inevitably achieves unsatisfying performance training in our dateset with complicated background and  complicated poses. 

\begin{figure}
  \centering
  
  \setlength{\abovecaptionskip}{0.cm}
   \includegraphics[width=1.0\linewidth]{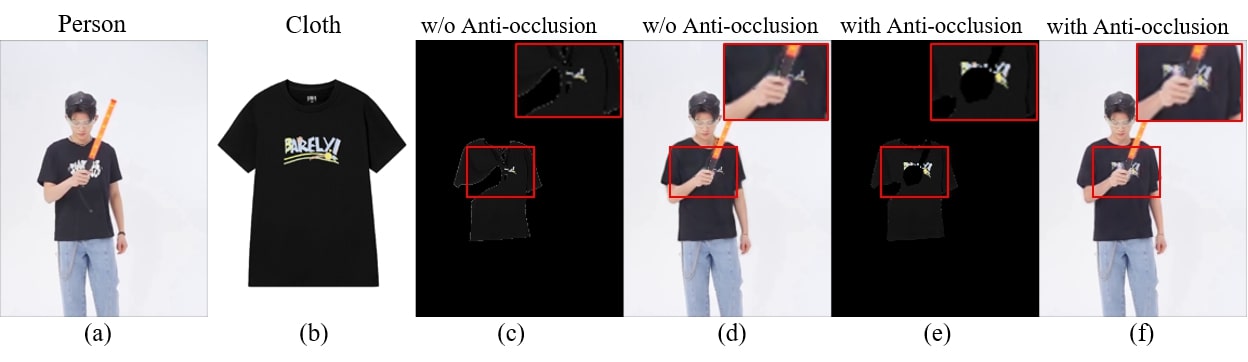}

   \caption{Effects of Anti-occlusion Warp Module.}
   \label{anti1}
   \vspace{-0.2cm}
\end{figure}

\begin{figure}
  \centering
  \setlength{\abovecaptionskip}{0.cm}
   \includegraphics[width=1.0\linewidth]{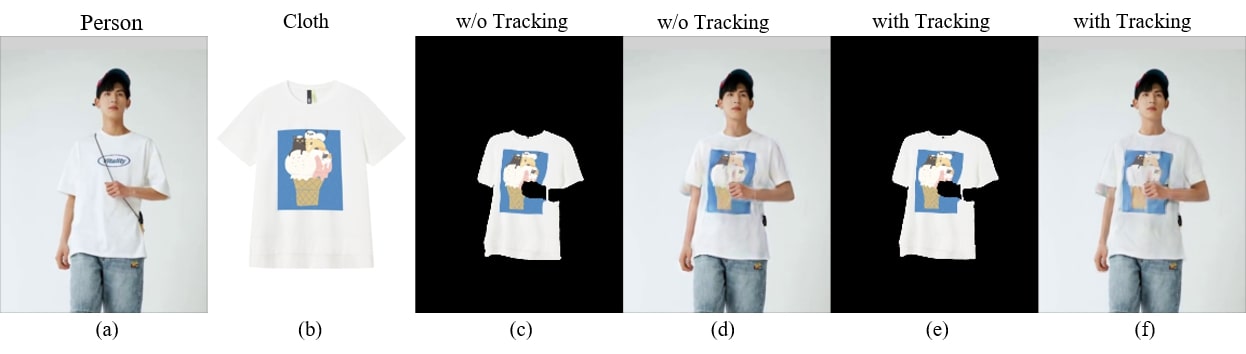}

   \caption{Effects of the Appearance-flow Tracking Module.}
   \label{tracking1}
   \vspace{-0.2cm}
\end{figure}

\begin{figure}
  \centering
  \setlength{\abovecaptionskip}{0.cm}
   \includegraphics[width=1.0\linewidth]{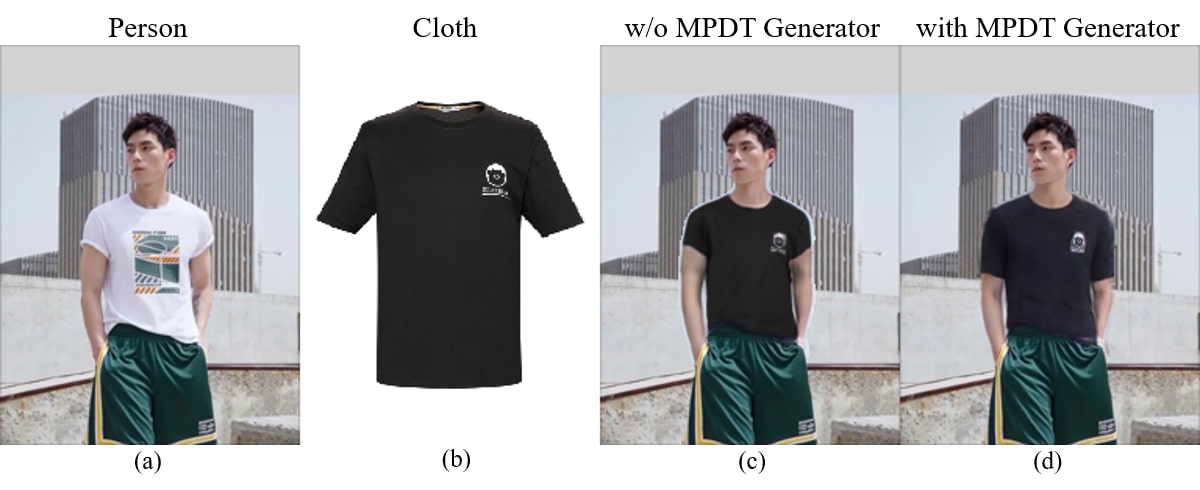}

   \caption{Effects of MPDT generator. }
   \label{former1}
   \vspace{-0.3cm}
\end{figure}

\subsection{Quantitative Analysis}\label{QUA} 
As shown in \cref{tab:example}, we perform the quantitative experiments in terms of both image-based evaluation metrics and video-based evaluation metrics. For image results, We use the structural similarity (SSIM)~\cite{ssim} and the learned perceptual image patch similarity (LPIPS)~\cite{liplis} to evaluate our method in the paired setting. For video results, we use the Video Frechet Inception Distance (VFID) to measure visual quality and temporal consistency in the unpaired setting, and both temporal and spatial features are extracted by two pre-trained video recognition CNN backbones: I3D~\cite{I3D} and 3D-ResNeXt101~\cite{101}. 

Compared with image-based methods and video-based methods on VVT dataset, ClothFormer outperforms them by a large margin, ClothFormer also surpasses the baseline methods with or without fusing the clothing-agnostic region on our new collected dataset, which demonstrates ClothFormer has great advantage in generating high-quality and spatio-temporally consistent try-on videos.

In pursuit of speed improvement, we try to compress ClothFormer by reducing the number of channels of MPDT from 256 to 96 and stacks of blocks from 8 to 6(ClothFormer-tiny), the FLOPs are reduced from 70.29G to 10.63G. However, the quantitative metrics are still better than other methods.
\subsection{Ablation Study}\label{AS}

We conduct an ablation study to analyze the designed modules of our method by creating three variants in our collected dataset, the Anti-occlusion Warp Module, Appearance-flow Tracking module and MPDT generator. 

\paragraph{Effectiveness of Anti-occlusion Warp Module.} As shown in \cref{anti1}, the ClothFormer$\dagger$ without Anti-occlusion Warp Module generates artifacts around occlusion with Pixel-squeeze phenomenon in the red box region of both warped clothes sequence and generated results. In contrast, our ClothFormer could generate satisfactory results, which demonstrates that the Anti-occlusion Warp Module has the capability to generate accurate warping against occlusion. 

\paragraph{Effectiveness of Appearance-flow Tracking.} \cref{tracking1} shows the texture of the clothes generated by ClothFormer$\divideontimes$ without Appearance-flow Tracking Module is flickering irregularly, verify Appearance-flow Tracking is beneficial to module produces a more temporally smooth result. 

\paragraph{Effectiveness of MPDT Generator.} As shown in \cref{former1}, the result of ClothFormer$\diamond$ with a U-Net try-on generator shows some weaknesses, such as, the non-target body parts looks fake, the region around the upper-body are blurring and the generated clothes are not harmony with the background. On the contrary, our ClothFormer with MPDT generator is capable to synthesize more video-realistic, natural and pleasing results, which demonstrates the superiority of dual-stream structure in the MPDT block.

In addition, as shown in \cref{tab:example}, the VFID scores of ClothFormer$\dagger$ are close to ClothFormer while there are gaps in SSIM and LPIPS, on the contrary, the values of SSIM and LPIPS of ClothFormer$\divideontimes$ are almost identical with ColthFormer while ClothFormer outperforms on VFID, which demonstrates that the Anti-occlusion Warp Module is beneficial to generate more accurate warping result to synthesize more photo-realistic images while the Appearance-flow Tracking Module helps to generate more  temporally smooth videos. And \cref{tab:example} also shows ClothFormer outperforms ClothFormer$\diamond$ on all indexes, which demonstrates the effectiveness of our MPDT generator.

\section{Conclusions}
We propose a novel video virtual try-on framework, \ie ClothFormer, which aims at generating realistic try-on videos while preserving the character of clothes, details of human identity (posture, body parts, bottom clothes) and background. We present three carefully designed modules, \ie Frame-level Anti-occlusion Warp module, Appearance-flow Tracking module and MPDT generator. Qualitative and quantitative experiments demonstrate that ClothFormer surpasses existing virtual try-on methods with a large margin.

\section*{Acknowledgements}
We are grateful to Qinkai Zheng for helping us revise the paper, and we thank Peipei Shi and Zhiqiang Qiao for their help in dataset collection.

{\small
\bibliographystyle{ieee_fullname}
\bibliography{egbib}
}
\appendix
{\noindent\Large\textbf{Supplementary Material}}
\newline
\setcounter{page}{1}
\input{final_supp.tex}

\end{document}

%% file: final_supp.tex
\thispagestyle{empty}
\appendix

\section{Implementation Details}
\subsection{Model Architectures}

\paragraph{GMM(TPS-based warping module).} To infer an anti-occlusion target clothes $\hat{C}_{t}$, we predict a TPS-based warped result by GMM, which is composed of two feature extractors and a regression network in GMM. First, we calculate a correlation matrix from two extracted features, and then predict TPS parameter $\theta$ by the regression network. There are 6 convolational layers in each feature extractor, and the regression network is composed of 4 convolution layers and a full-connection layer. The details of the GMM network are shown in \cref{GMM}.

\paragraph{AFWM(Appearance-flow-based warping module).} Similar to~\cite{PFAFN}, we set a dual pyramid network to extract features of $(A_t,D_t,P_t)$ and $\hat{C}_t$ in different scales $(c_N, b_N)$. For each scale, we adopt a Flow Network(FN) to estimate the dense appearance flow $f_n$. All FNs are cascaded to predict the finest flow $f_N$. In detail, the inputs of $n-{th}$ FN are $(c_n, b_n)$ and $f_{n-1}$, and the output is $f_n$.

\begin{figure}[b]
    \centering
    \setlength{\abovecaptionskip}{0.cm}
    \includegraphics[width=1.0\linewidth]{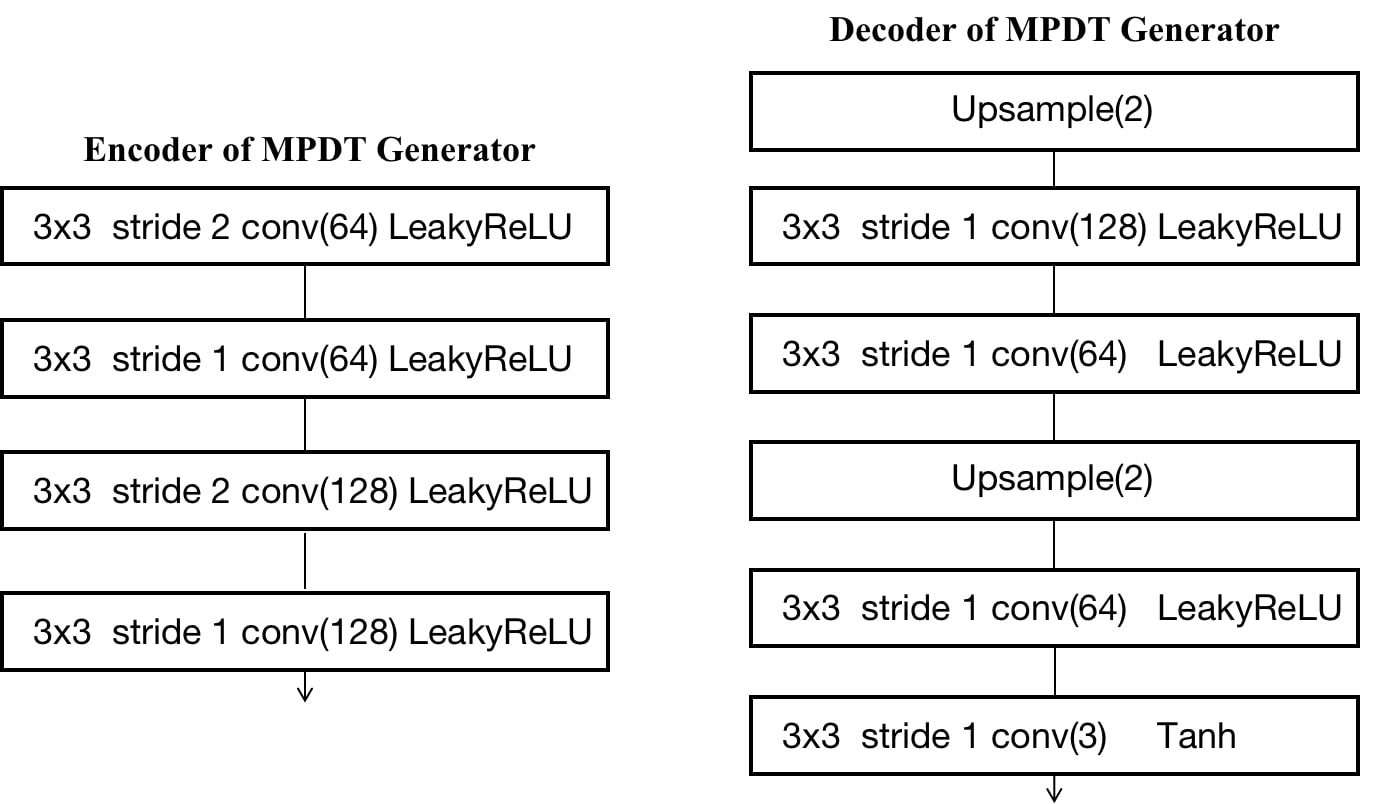}
    \caption{Encoder and Decoder architecture of MPDT Generator}
    \label{encoder}
    \vspace{-0.2cm}
\end{figure}

\paragraph{Encoder and Decoder in MPDT Generator.} As described in the main text, there are three identical frame-level encoders and one frame-level decoder. There are 4 convolution layers with two times downsampling in encoder, and 4 convolution layers and 2 upsampling layers in decoder. The details of encoder and decoder are shown in \cref{encoder}.

\subsection{Training Details}
\paragraph{Optimization.} We use Adam~\cite{adam} as optimizer with $ \beta _{1} = 0.5 $, $ \beta _{2} = 0.999 $, a fixed learning rate of 0.0002 and all with a batch size of 16.

\paragraph{Hardware.} All the codes are implemented by deep learning toolkit PyTorch and 8 NVIDIA V100 GPUs are used in our experiments. The training takes around 2 days for TPS-based warping module in Frame-level Anti-occlusion Warp Module, and around 3 days for Appearance-flow warping module in Frame-level Anti-occlusion Warp Module, and around 4 days for MPDT Generator. 
 
\paragraph{Loss detail in Anti-occlusion Warp Module.} The full loss $L^{TPS\textrm{-warp}}_{t}$ for TPS-based warping mudule are written as Eq. (1), we set the hyper-parameter $ \lambda^{s d c}_{t} $ to 0.04 and the second-order difference constraint $ L^{s d c}_{t} $ detailed as:
 
 \begin{equation}
 \begin{aligned}
   L^{s d c}_{t}&=\sum\limits_{p \in P}\left| {\left \| pp_{0} \right\|}_{2} - {\left \| pp_{1} \right\|}_{2} \right|+\left| {\left \| pp_{2} \right\|}_{2} - {\left \| pp_{3} \right\|}_{2} \right|
   \\
   + &\left \| S(pp_{0}-pp_{1})\right\|+\left\| S(pp_{2}-pp_{3})\right\|
  \end{aligned}
   \tag{13}
  \label{eq:important_supp}
\end{equation}

where symbol $p$ denotes a certain sampled TPS control point and $p_{0}$, $p_{1}$, $p_{2}$ and $p_{3}$ are top, bottom, left and right point of $p$, respectively. The $S(p, pi)$ is the slope between $p$ and $p_{i}$. 
 
  
  The full loss $L^{flow-\text { warp }}_{t}$ for Appearance-flow-based warping mudule are written as Eq. (2), we set the hyper-parameter $ \lambda^{sec}_{t} $ to 20 and the second-order smooth constraint $ L^{sec}_{t} $ detailed as:
 \begin{equation}
 \begin{aligned}
   L^{sec}_{t}&=\sum\limits_{i=1}^{N}\sum\limits_{t}\sum\limits_{\pi \in N_{t}}P(f_{i}^{t-\pi}+f_{i}^{t+\pi}-2f_{i}^{t})
  \end{aligned}
   \tag{14}
  \label{sec}
\end{equation}

Where $P$ denotes the generalized charbonnier loss function~\cite{lll1}, $f_{i}^{t}$ is the $t^{th}$ point on the appearance-flow maps of $i^{th}$ scale. $N_{t}$ indicates the set of horizontal, vertical, and both diagonal neighborhoods around the $t^{th}$ point. 
 \begin{figure}
\setcounter{figure}{8}
    \centering
    \setlength{\abovecaptionskip}{0.cm}
    \includegraphics[width=1.0\linewidth]{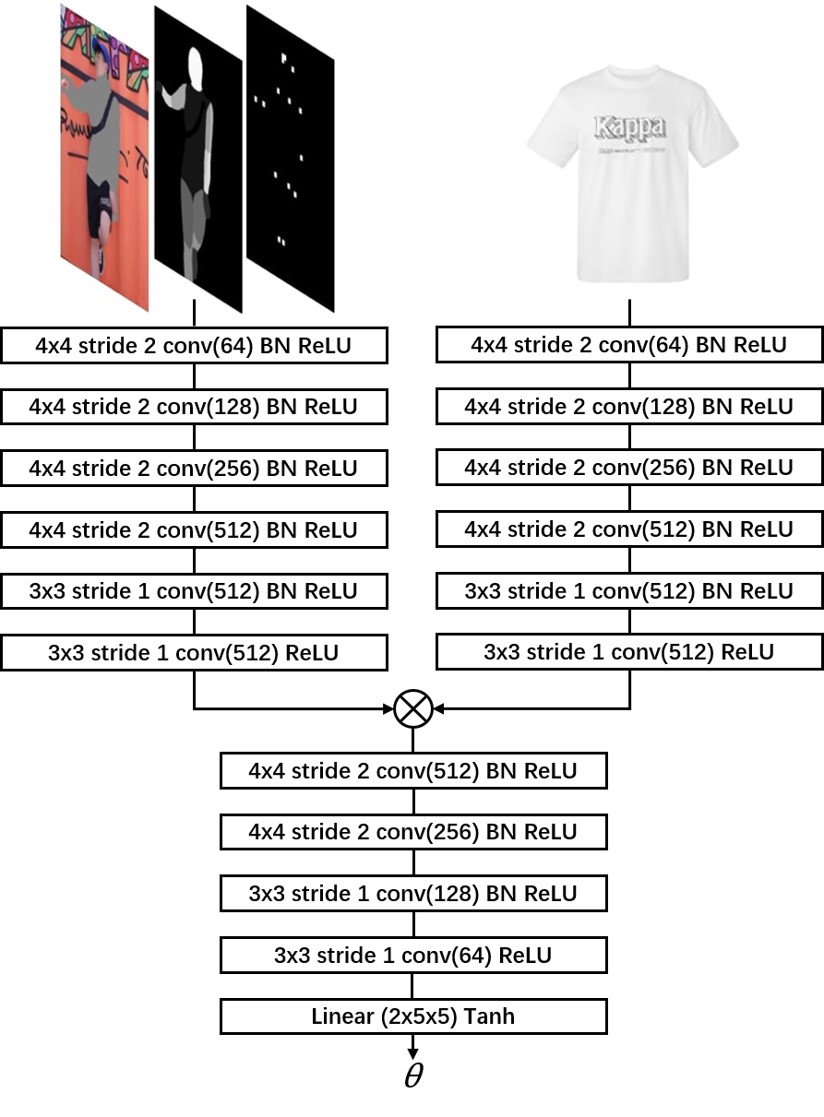}
    \caption{GMM network}
    \label{GMM}
    \vspace{-0.5cm}
    \setlength{\belowcaptionskip}{-1cm}
\end{figure}

\paragraph{Matrix X in Appearance-flow Tracking Module.} We use ridge regression~\cite{oridge} to track appearance-flow to obtain temporally smooth warped clothing sequences as shown in Eq. (3), the feature matrix X detail as: 
$$
X = \left[
\begin{matrix}
 x_{t-N}^{(1)}      & x_{t-N+1}^{(1)}      & \cdots & x_{t-1}^{(1)}      \\
 y_{t-N}^{(1)}      & y_{t-N+1}^{(1)}      & \cdots & y_{t-1}^{(1)}      \\
 x_{t-N}^{(2)}      & x_{t-N+1}^{(2)}      & \cdots & x_{t-1}^{(2)}      \\
 y_{t-N}^{(2)}      & y_{t-N+1}^{(2)}      & \cdots & y_{t-1}^{(2)}      \\
 \vdots & \vdots & \ddots & \vdots \\
 x_{t-N}^{(W \times H)}      & x_{t-N+1}^{(W \times H)}      & \cdots & x_{t-1}^{(W \times H)}      \\
 y_{t-N}^{(W \times H)}      & y_{t-N+1}^{(W \times H)}      & \cdots & y_{t-1}^{(W \times H)}      \\
\end{matrix}
\right]
$$

Here, $ x_{m}^{n} $ and $ y_{m}^{n} $ (for m = 1...N; n = 1...$ W \times H $) are the
directly estimated horizontal and vertical coordinate values for $ n \_ th $ point in appearance-flow at frame $m$, and we set N=3 for high computational efficiency. After Eq. (3) we reshape $ \hat{f}_{t}^{1D} $ back to $ \hat{f}_{t} $  with original spatial size $ W \times H $.

\paragraph{Loss detail in MPDT Generator.} The Full loss $L_{\textrm {try-on}}$ for MPDT generator are written as Eq. (12), we set the hyper-parameter $\lambda_{1} = \lambda_{3} = 1, \lambda_{4} = 0.01, \lambda_{2} = 10$.
And $L_{l1}^{clothes} $ is the L1 loss in clothing regions and denoted as:
 \begin{equation}
 \begin{aligned}
   	L_{l1}^{clothes} = \frac{\left \| {M_{C}}_{1}^{T}\odot(I_{1}^{T}-\Tilde{I}_{1}^{T}) \right\|}{\left \| {M_{C}}_{1}^{T} \right\|}
  \end{aligned}
   \tag{15}
  \label{L1_cloth}
\end{equation}
where $\odot$ indicates element-wise multiplication.

The perceptual loss function $ L_{\textrm {perc}}$ is denoted as:
 \begin{equation}
 \begin{aligned}
   	L_{\textrm {perc}} = \sum\limits_{i=1}^{N_{l}} \frac{1}{Ne_{i}}[\left \| F^{(i)}(I_{1}^{T})-F^{(i)}(\Tilde{I}_{1}^{T})\right \|]
  \end{aligned}
   \tag{16}
  \label{perc}
\end{equation}
where $N_{l}$ is the number of features extracted from different layers of the VGG network $F$~\cite{perceptual}, and $F^{(i)}$ and $R_{i}$ are the activation and the number of elements in the $i^{th}$ layer of F, respectively. 

The adversarial loss $ L_{\textrm{TPGAN}} $ is denoted as:
 \begin{equation}
 \begin{aligned}
   	L_{\textrm{TPGAN}} = -E_{z\backsim {p}_{\Tilde{I}_{1}^{T}}}(z)[D(z)]
  \end{aligned}
   \tag{17}
  \label{gan}
\end{equation}

And the optimization function for the T-PatchGAN discriminator is shown as follows:
 \begin{equation}
 \begin{aligned}
   	L_{\textrm{TPGAN}} = E_{x\backsim {p}_{{I}_{1}^{T}}}(x)[RELU(1-D(x))]\\
   	+E_{z\backsim {p}_{\Tilde{I}_{1}^{T}}}(z)[RELU(1+D(z))]
  \end{aligned}
   \tag{18}
  \label{d}
\end{equation}

\section{Additional Experiments}
\subsection{Comparison with different warping methods}
\begin{figure*}[htp]
    \centering
    \setlength{\abovecaptionskip}{0.cm}
    \includegraphics[width=1.0\linewidth]{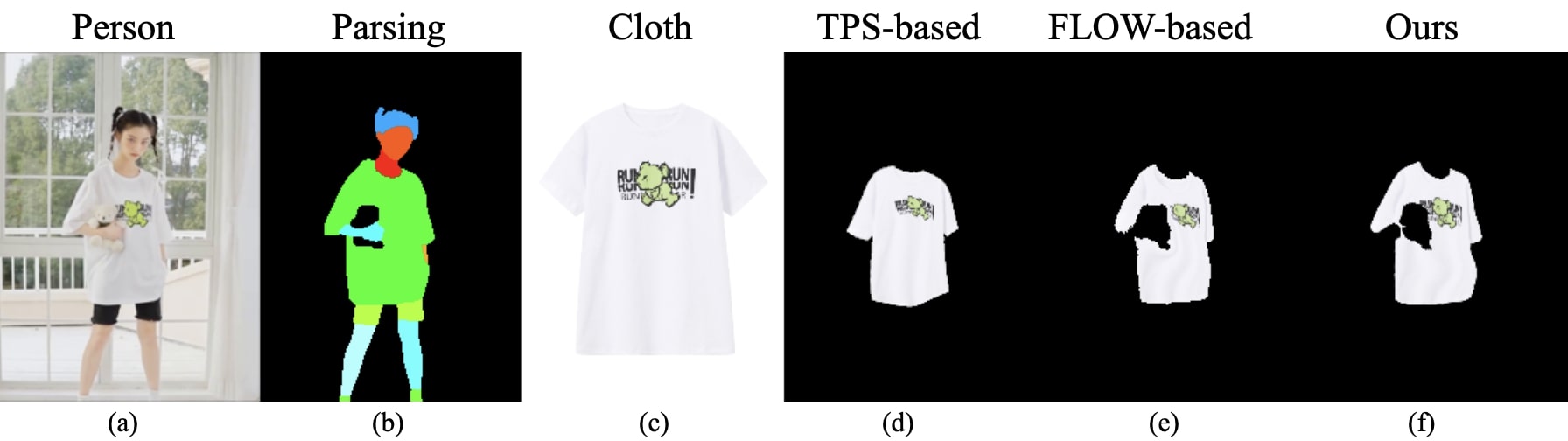}
    \caption{Qualitative comparison of warping methods.}
    \label{warp_res}
    \vspace{-0.2cm}
\end{figure*}

To demonstrate TPS-based warping can handle partial occlusions but lead to the misalignment between the warped clothes and the appearance-flow-based methods predicts more accurate deformations but are very sensitive to occlusions, we conduct a comparison experiment as shown in \cref{warp_res}. Obviously, the shape of the clothes warped by TPS-based warping method is different from the shape of the clothes worn on the reference person. On the contrary, the shape of the clothes warped by the appearance-flow-based is better but the pixel-squeeze phenomenon appears around the doll. Compared to the baseline method, our two-stage anti-occlusion warping method warped a both accurate and anti-occlusion results. 
\subsection{Baseline methods without fusing background}
\begin{figure*}[htp]
    \centering
    \setlength{\abovecaptionskip}{0.cm}
    \includegraphics[width=1.0\linewidth]{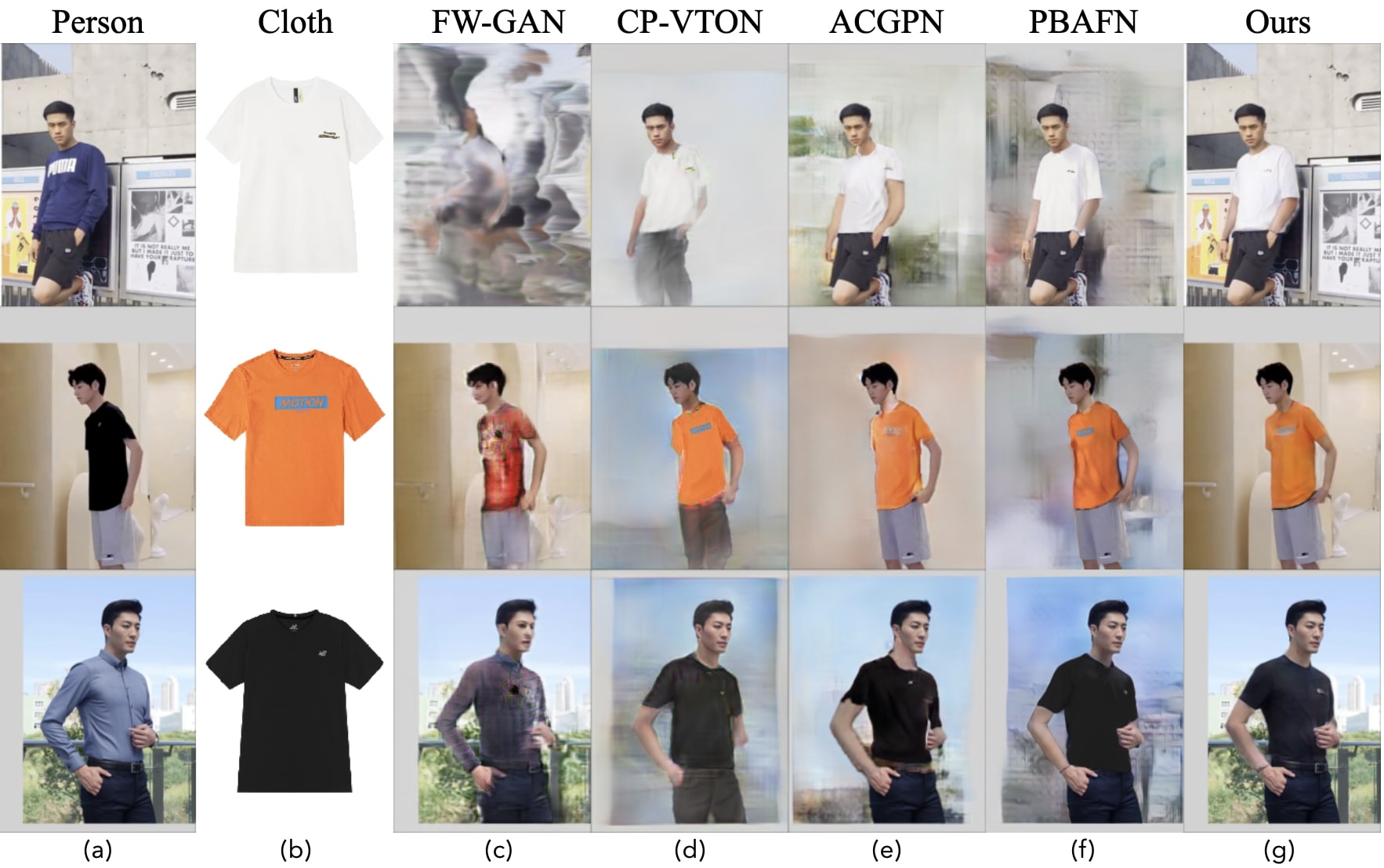}
    \caption{Qualitative comparison of the baseline methods w/o fusing background.}
    \label{all_method_w_bg}
    \vspace{-0.2cm}
\end{figure*}

As shown in \cref{all_method_w_bg}, the baseline methods without fusing background fail to reconstruct complex background when training and testing on our collected wild virtual try-on dataset.  

\subsection{Qualitative Results}
\begin{figure*}[htp]
    \centering
    \setlength{\abovecaptionskip}{0.cm}
    \includegraphics[width=0.9\linewidth]{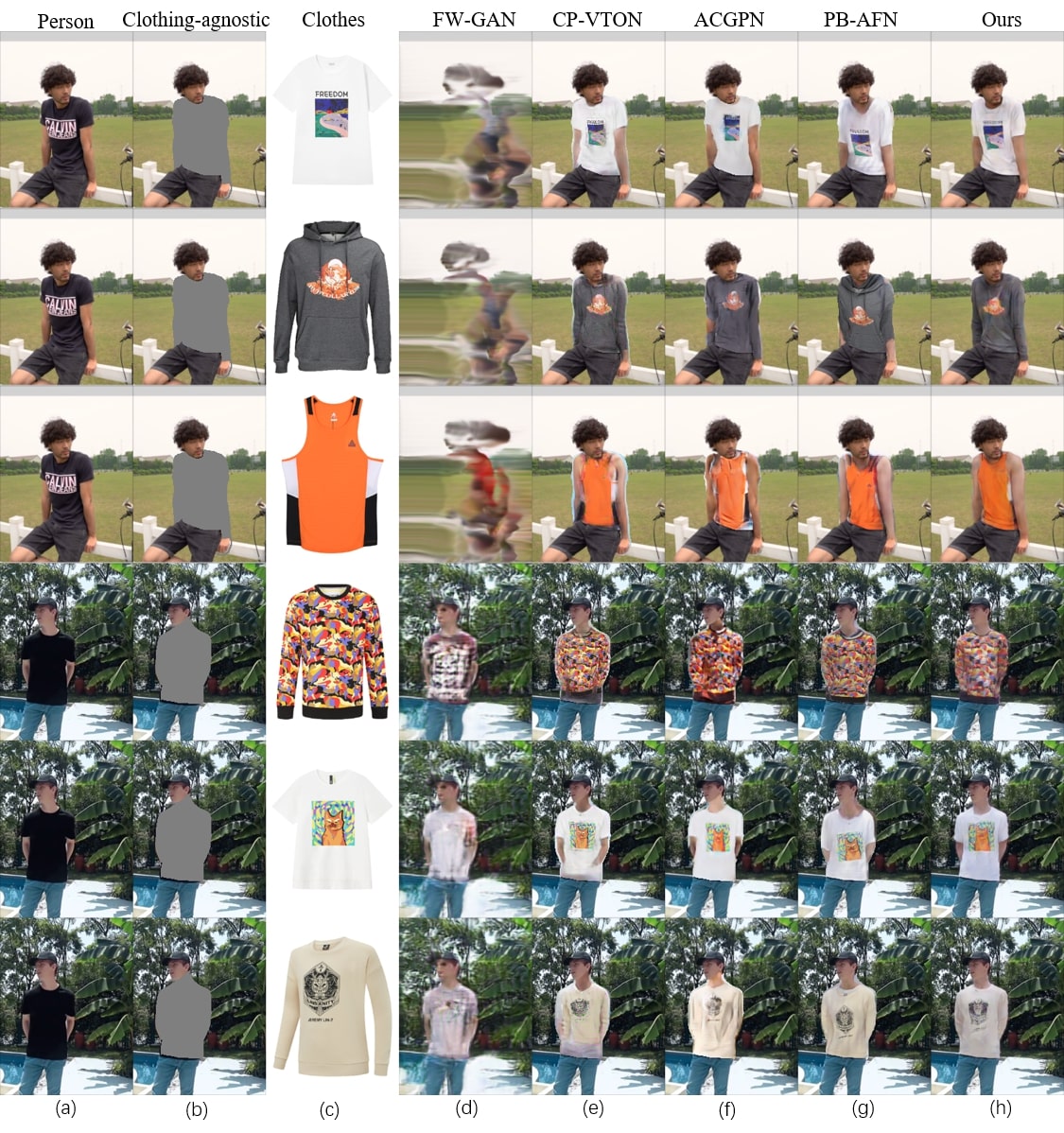}
    \caption{Qualitative comparison of the baseline methods.}
    \label{compare}
    \vspace{-0.2cm}
\end{figure*}

\begin{figure*}[htp]
    \centering
    \setlength{\abovecaptionskip}{0.cm}
    \includegraphics[width=0.9\linewidth]{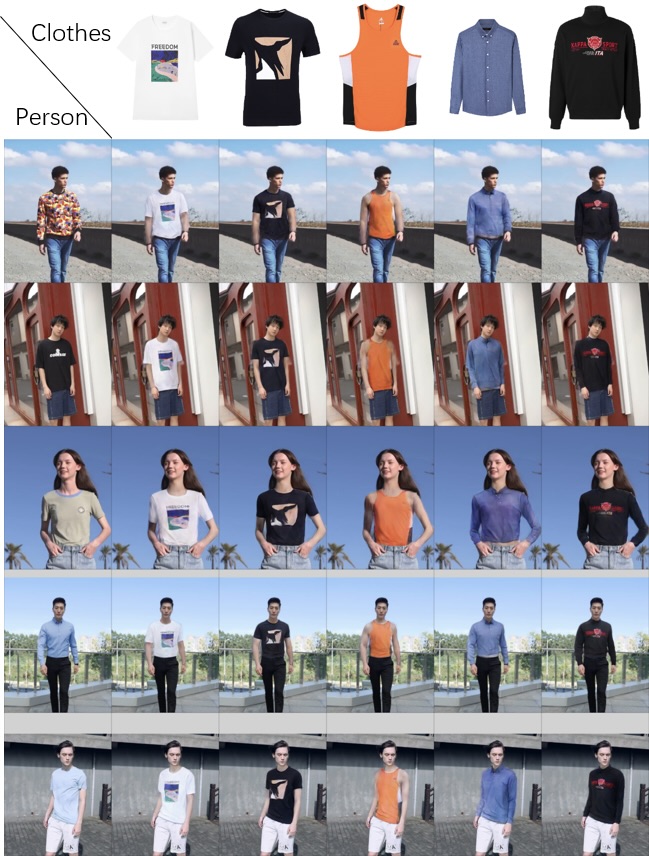}
    \caption{Additional qualitative results of ClothFormer on our dataset.}
    \label{show}
    \vspace{-0.2cm}
\end{figure*}

\begin{figure*}[htp]
    \centering
    \setlength{\abovecaptionskip}{0.cm}
    \includegraphics[width=0.9\linewidth]{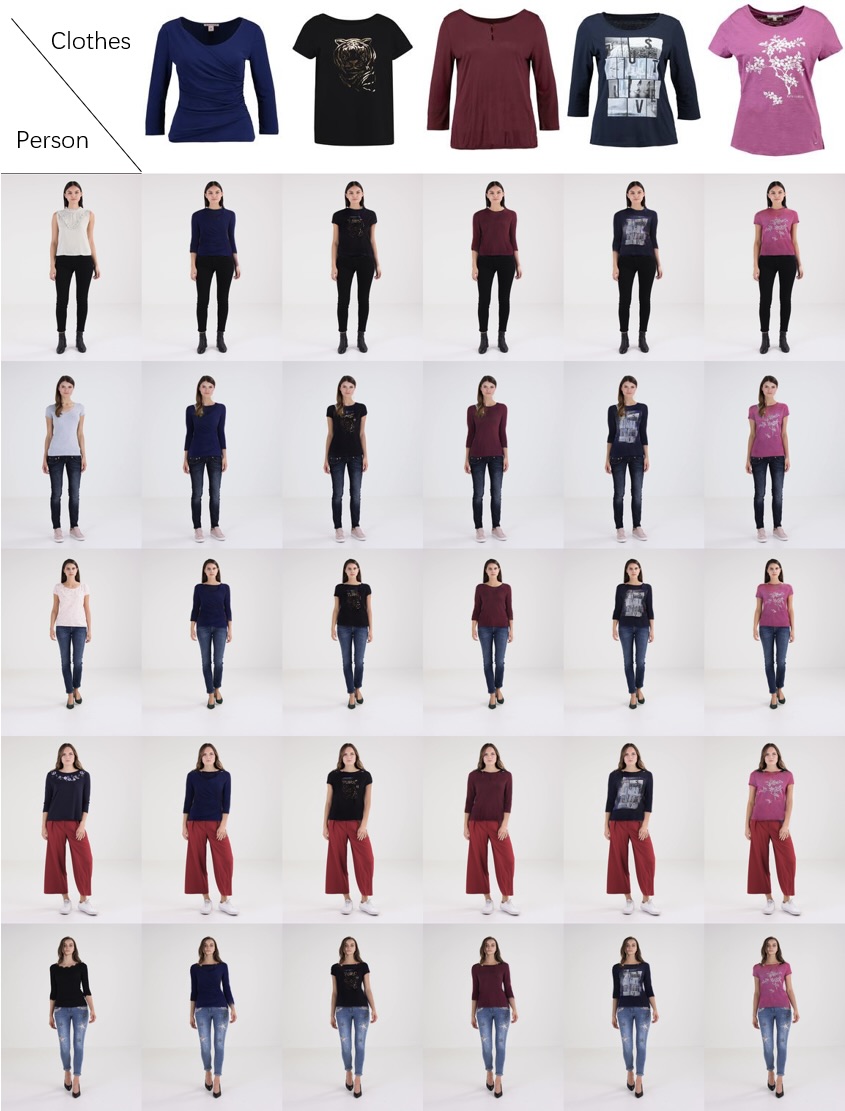}
    \caption{Additional qualitative results of ClothFormer on the VVT dataset.}
    \label{show_vvt}
    \vspace{-0.2cm}
\end{figure*}

We provide additional qualitative results to demonstrate our model’s capability of generating a temporally smooth and photo-realistic video. \cref{compare} show the qualitative comparison of the baselines in our wild virtual try-on dataset. \cref{show} and \cref{show_vvt} show additional results of ClothFormer in our dataset and the VVT dataset, respectively.

\section{Failure Cases and Limitations}

\begin{figure*}[thbp]
    \centering
    \includegraphics[width=1.0\textwidth]{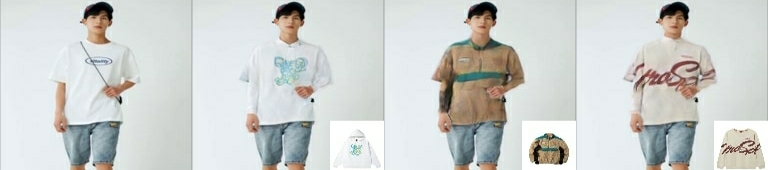}
    \makebox[1.0\textwidth]{\small (a) Failure cases of redundant sleeve.}
    \\ \vspace{0.5em}
    \includegraphics[width=1.0\textwidth]{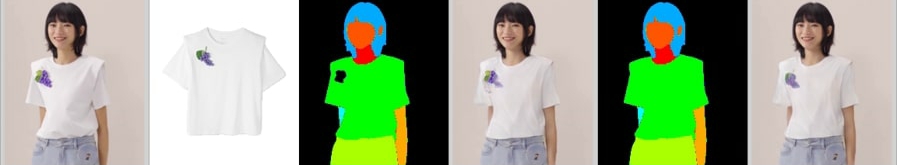}
    \makebox[1.0\textwidth]{\small (b) Failure cases of inaccurate parsing results.}
    \caption{Failure cases and Limitations}
    \label{failure_cases}
\end{figure*}

As shown in the forth column of \cref{failure_cases}b, similar to existing parser-based methods, when parsing results are inaccurate (the grape pattern on the clothes is predicted as occlusion), ClothFormer generates visually terrible try-on images with noticeable artifacts in the inaccurate region. However, as shown in the sixth column of \cref{failure_cases}b, ClothFormer can produce a realistic and natural video result when we fix the parsing results manually, which shows it can be a valuable future direction to generate try-on videos by developing a parser-free video virtual try-on method. As shown in \cref{failure_cases}a, ClothFormer generates a redundant sleeve when short-sleeve clothes to long-sleeve clothes, which might be solved by predicting a human-parsing maps of a person wearing the target clothes to guide the try-on video synthesis like ACGPN~\cite{ACGPN} and VITON-HD~\cite{VITON-HD} does, but we need to predict a temporally smooth parsing sequence to avoid flickering results in time dimension.

One of the limitations of our model is that ClothFormer is not able to generate clothes with textured patterns on the back, instead, ClothFormer learns a pure color on the back due to most clothes are pure color on the back in both VVT \newpage dataset and our dataset. The root cause of the above problem is that only the front of the target clothes are available, we believe that ClothFormer has the capability to cover the case when training with both front of the target clothes and back of the target clothes as input.

\clearpage
\clearpage

 